\documentclass[letterpaper, 10 pt, conference]{ieeeconf}  % Comment this line out if you need a4paper

\IEEEoverridecommandlockouts                              % This command is only needed if 
\usepackage{graphics} % for pdf, bitmapped graphics files
\usepackage{graphicx}
\usepackage{epsfig} % for postscript graphics files
\usepackage{mathptmx} % assumes new font selection scheme installed
\usepackage{times} % assumes new font selection scheme installed
\usepackage{amsmath} % assumes amsmath package installed
\usepackage{amssymb}  % assumes amsmath package installed
\usepackage{float}
\usepackage{tikz}

\usepackage{threeparttable}

\title{\LARGE \bf
A Novel Twisted-Winching String Actuator for Robotic Applications: Design and Validation
}
\author{Ryan Poon*, Vineet Padia*, and Ian W. Hunter % <-this % stops a space
\thanks{*These authors contributed equally to this work.}% <-this % stops a space
\thanks{The authors are with the BioInstrumentation Laboratory, Massachusetts Institute of Technology, Cambridge, MA 02139, United States of America (email: rpoon@mit.edu; vpadia@mit.edu; ihunter@mit.edu)}%
}

\begin{document}

\maketitle
\thispagestyle{empty}
\pagestyle{empty}

%%%%%%%%%%%%%%%%%%%%%%%%%%%%%%%%%%%%%%%%%%%%%%%%%%%%%%%%%%%%%%%%%%%%%%%%%%%%%%%%
\begin{abstract}

This paper presents a novel actuator system combining a twisted string actuator (TSA) with a winch mechanism. Relative to traditional hydraulic and pneumatic systems in robotics, TSAs are compact and lightweight but face limitations in stroke length and force-transmission ratios. Our integrated TSA-winch system overcomes these constraints by providing variable transmission ratios through dynamic adjustment. It increases actuator stroke by winching instead of overtwisting, and it improves force output by twisting. The design features a rotating turret that houses a winch, which is mounted on a bevel gear assembly driven by a through-hole drive shaft. Mathematical models are developed for the combined displacement and velocity control of this system. Experimental validation demonstrates the actuator's ability to achieve a wide range of transmission ratios and precise movement control. We present performance data on movement precision and generated forces, discussing the results in the context of existing literature. This research contributes to the development of more versatile and efficient actuation systems for advanced robotic applications and improved automation solutions.

\end{abstract}

%%%%%%%%%%%%%%%%%%%%%%%%%%%%%%%%%%%%%%%%%%%%%%%%%%%%%%%%%%%%%%%%%%%%%%%%%%%%%%%%
\section{INTRODUCTION}

Twisted string actuators (TSAs) offer a compact, lightweight alternative to traditional hydraulic and pneumatic systems in robotics. TSAs convert rotational motion into linear motion by twisting high-strength strings, achieving high transmission ratios without complex mechanisms or bulky gearboxes. These benefits are particularly evident in devices constrained by space, weight, and complexity, such as haptic gloves \cite{hosseini_exoten-glove_2018}, robotic hands \cite{nedelchev_design_2018}, exoskeletons \cite{kornbluh_twisted_2018, seong_development_2020}, and soft robots \cite{bombara_twisted_2021}. 

\begin{figure}[t]
    \centering
    \includegraphics[width = 0.99 \linewidth]{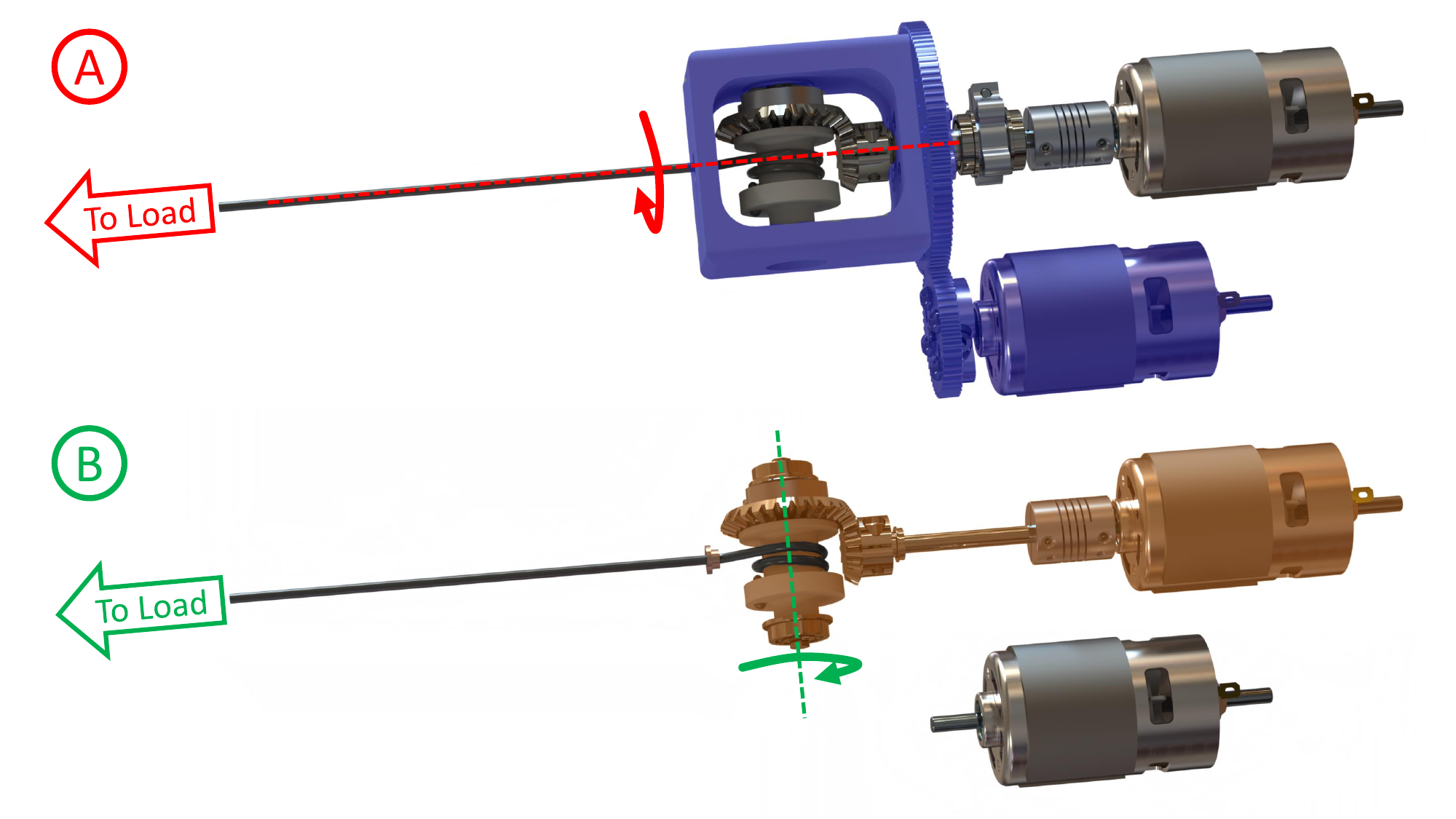}
    \caption{The TSA/winch hybrid actuator. (A) shows the actuator without all supporting brackets, standoffs, and fasteners, with the blue region highlighting the turret mechanism for twisting the string. (B) shows the same actuator without the turret gear transmission to better see the through-hole shaft that drives the bevel gear train and winch (highlighted orange).}
    \label{fig:actuator}
\end{figure}

Despite their simplicity, TSAs exhibit nonlinear behavior, prompting extensive efforts to model and characterize their properties. Past work has improved on earlier models, demonstrating new models' efficacy across a range of string materials and routing parameters \cite{gaponov_twisted_2013, palli_twisted_2016}. High-accuracy modeling of the dynamics that considers friction and string compliance has also been achieved \cite{nedelchev_accurate_2020}. Several control strategies have been developed to manage this behavior \cite{nedelchev_high-bandwidth_2019, rodriguez_hybrid_2020, palli_modeling_2012, baek_tension_2023, wurtz_twisted_2010}, and novel measurement methods to handle noise and parameter uncertainties for control techniques have also been studied \cite{lin_novel_2018}. Even the lifespan of TSAs has been explored \cite{usman_study_2017}.

These studies, however, highlight some drawbacks of TSAs, leading to various modifications to address their limitations. Because the range of transmission ratios can be limited, some researchers have implemented discretely adjustable systems like clutches \cite{jeong_2-speed_2018} or manually modifiable modules \cite{singh_passively_2015} to switch between high-speed and high-force modes. Others have pursued continuously variable transmission through passive elastomeric mechanisms \cite{kim_elastomeric_2020} or by controlling the rotation ratio of individual string twists \cite{jang_active-type_2022}. Yet, these solutions do not fully address the challenges brought upon by the low stroke linear movement typical of TSAs, its proportional relationship to force transmission, and the limitations due to overtwisting. Some have tried parallel TSAs to increase the force-to-stroke ratio at the cost of adding motors, while others have made compliant thermally activated conductive polymers to enable large strain TSAs \cite{bombara_compliant_2022}. Optimization techniques have also been applied to maximize the stroke relative to the original string length \cite{baek_enhancing_2024, tavakoli_compact_2016}. Despite these advancements, these TSA systems encounter the same issue—excessive rotations lead to overtwisting, limiting the actuator's overall stroke unless the unstable overtwisting regime is closely modeled and exploited \cite{tavakoli_compact_2016, konda_experimental_2023}.

\begin{table}
\centering
\begin{threeparttable}
\caption{Advantages and Disadvantages}
\begin{tabular}{|c|c|c|c|}
\hline
\textbf{Parameters} & \textbf{TSA} & \textbf{Winch} & \textbf{Hybrid} \\
\hline
Weight/Size & $+$ & $-*$ & $-$ \\
Stroke Length & $-$ & $+$ & $+$ \\
Force Output & $+$ & $-$ & $+$ \\
\hline
\end{tabular}
\begin{tablenotes}
\item[*] Depends if there is a gearbox present.
\end{tablenotes}
\end{threeparttable}
\end{table}

\begin{figure*}[t]
    \centering
    \includegraphics[trim=0pt 80pt 0pt 80pt, clip, width = 0.99 \linewidth]{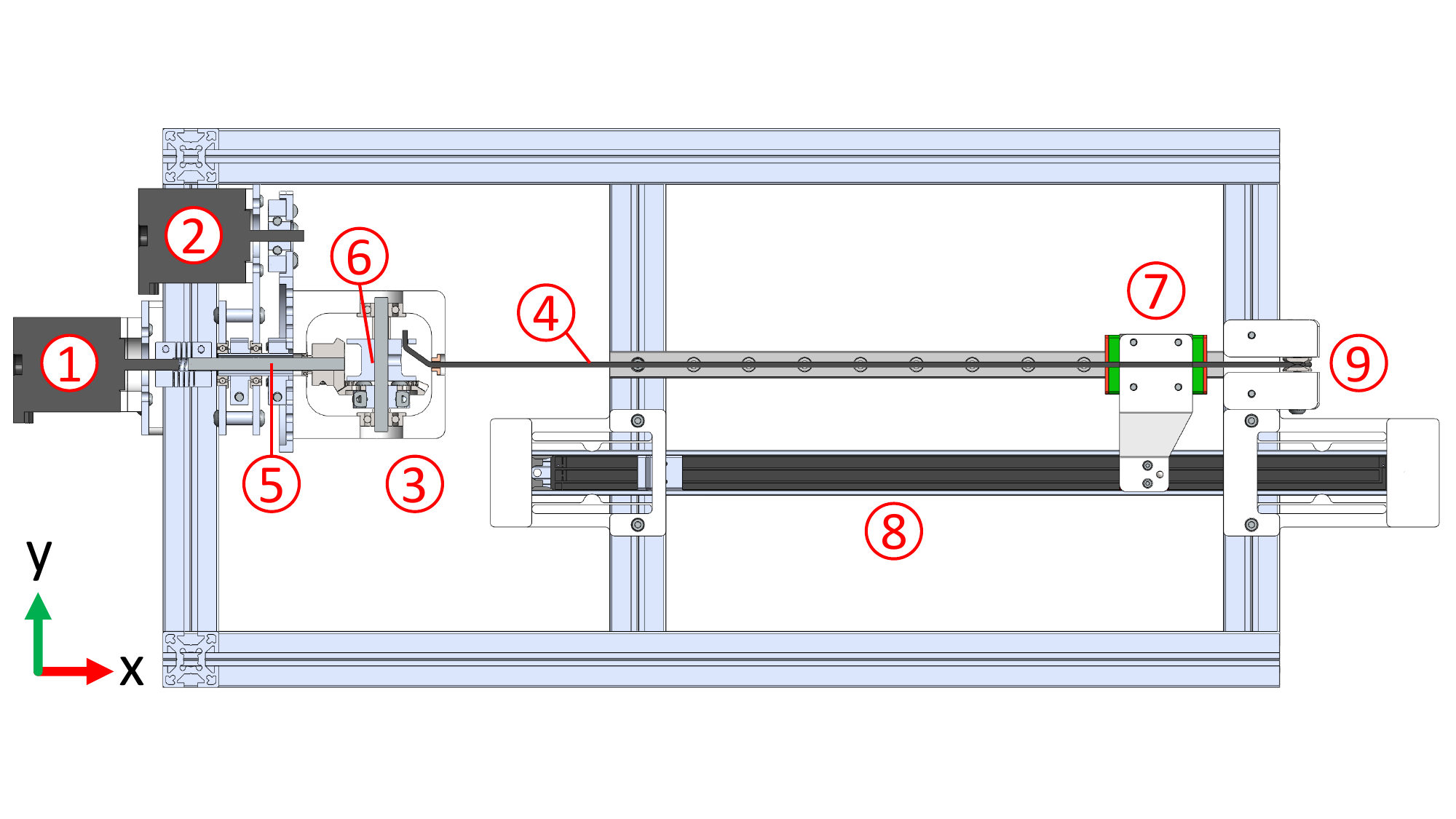}
    \caption{Cross section of the actuator and experimental setup. Motors (1) and (2) drive the winch and the turret, respectively. Motor 2 rotates a turret (3) via gears to twist the string (4). Motor 1 drives a drive shaft (5) that goes through the gears to rotate bevel gears which in turn rotates the winch (6). The string pulls a linear carriage (7), the position of which is measured by a linear potentiometer (8). The string is redirected over the edge of a table at (9) to a hanging adjustable set of masses that acts as a load force. For the force-torque experiments, (9) is replaced with a load cell that the string is fixed to.}
    \label{fig:crosssection}
\end{figure*}

This study introduces a novel mechanism combining a TSA with a winch system to overcome these limitations (Figs. \ref{fig:actuator} and \ref{fig:crosssection}). Recent modeling efforts confirm that while TSAs excel in high force output relative to torque input, winches perform best when large contraction ranges and speeds are required \cite{tsabedze_model-based_2024, tsabedze_compact_2021}. However, to the authors' knowledge, no system has combined the advantages of both. The integrated TSA-winch system provides variable transmission ratios through dynamic adjustment, enhancing actuator stroke and force flexibility for precise, adaptable movement without overtwisting. The performance of this system was evaluated using a rotating turret design that houses a winch mounted on a bevel gear assembly driven by a through-hole drive shaft. Although this design requires one extra motor, the flexibility in motor size and absence of heavy on-axis gearboxes suggest good potential for miniaturization while maintaining large stroke and transmission ratio variability.

To show these advantages, this paper first describes the modeling, experimental setup, and methodologies. It then presents performance data on movement precision and force outputs, followed by a discussion of the results. The conclusion summarizes the findings and their implications for robotics and automation. By enhancing TSA capabilities, this research aims to develop more efficient and versatile actuation systems for advanced robotic applications and improved automation solutions.

\section{MODELS AND EQUATIONS}

\begin{figure}[t]
    \centering
        \includegraphics[width = 0.99 \linewidth]{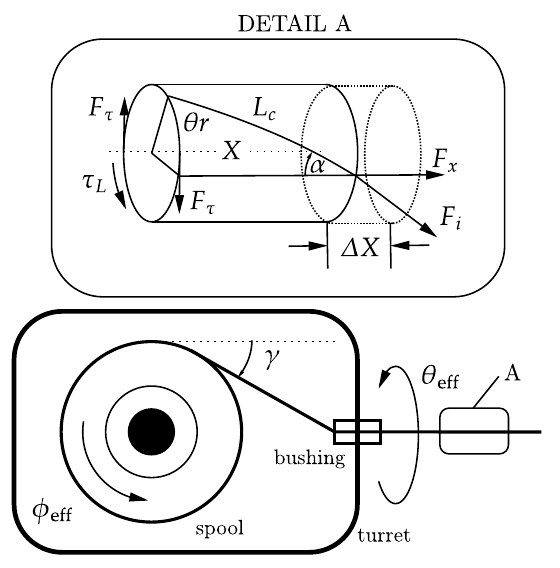}
    \caption{Line diagram showing all relevant variables.}
    \label{fig:linediagram}
\end{figure}

\subsection{Twisted String Actuation}
In this section, the baseline model for TSAs is based on those given in \cite{gaponov_twisted_2013}. The contracted length $X$ of a twisted loaded string is given by:

\begin{equation}
\label{eq:length}
X = \sqrt{L_c^2 - \theta^2 r_0^2}
\end{equation}
where $L_c$ is the extended length, $r_0$ is the initial radius of the string, and $\theta$ is the twisting angle (Fig. \ref{fig:linediagram}). If it is assumed that the radius of the string increases during twisting, then the linear contraction of the string $\Delta X$ is:

\begin{equation}
\label{eq:linearcontraction}
\Delta X = L_c - \sqrt{L_c^2 - \theta^2 r_{\text{var}}^2}
\end{equation}
where $r_{\text{var}} = r_0 \sqrt{\frac{L_c}{X}}$ accounts for varying radius of the string as it twists \cite{gaponov_twisted_2013}.

\subsection{Winch and Actuator Gear System}
For motors 1 and 2 as described in Fig. \ref{fig:crosssection} rotating by angles $\theta_1$ and $\theta_2$, the effective winch rotation $\phi_{\text{eff}}$ and twisting angle $\theta_{\text{eff}}$ are:

\begin{equation}
\label{eq:rotationandtwistangles}
\phi_{\text{eff}} = \frac{\theta_1(-1)^{a + 1} + \theta_\text{eff}}{N_\phi}, \quad \theta_{\text{eff}} = \frac{\theta_2}{N_\theta}(-1)^{b + 1}
\end{equation}
where the number of gears and the gear ratio of the bevel gear assembly housed in the turret are $a$ and $N_\phi$, and the number of gears and the gear ratio of the gear train driving the turret are $b$ and $N_\theta$. Note that the direction of rotation is based on the axes in Fig. \ref{fig:crosssection}. Equation (\ref{eq:rotationandtwistangles}) is general such that this actuator design can be altered for different geartrain assemblies. For example, to increase the winch size, more gears in the turret housing may be needed. Also note that winch rotation occurs when the turret rotates but motor 1 rotates at a different velocity, causing relative motion between the bevel gears in the turret housing.

From the desired twist and winch angles, the motors' angular positions can be deduced. The angular velocities are:

\begin{equation}
\label{eq:motorvelocities}
\dot{\theta}_1 = \frac {N_\phi \dot{\phi}_\text{eff} - \dot{\theta}_\text{eff}}{(-1)^{a + 1}}, \quad \dot{\theta}_2 = \frac{N_\theta \dot{\theta}_\text{eff}}{(-1)^{b + 1}}
\end{equation}

\subsection{Combined Displacement Model}
Equation (\ref{eq:length}) can be adjusted to account for both twisting and winching, giving the total linear contraction $\Delta X_{\text{total}}$ as:

\begin{equation}
\label{eq:linearcontractiontotal}
\Delta X_{\text{total}} = L_c - \sqrt{L_c^2 - \theta_{\text{eff}}^2 r_{\text{var,eff}}^2} + r_w \phi_\text{eff}
\end{equation}
where $r_w$ is the winch radius and $r_{\text{var,eff}} = r_0 \sqrt{\frac{L_c}{X_{\text{total}}}}$. The contracted length $L_c$ in Equations (\ref{eq:length}), (\ref{eq:linearcontraction}), and (\ref{eq:linearcontractiontotal}) depends on the axial force $F_x$ and the winch action:

\begin{equation}
\label{eq:forcesandcontractedlength}
L_c = L + \frac{F_i}{K} - r_w\phi_\text{eff}, \quad F_i = \sqrt{F_\tau^2 + F_x^2}, \quad F_\tau = \frac{\tau_L}{r_0}
\end{equation}
where $L$ is the unloaded string length, $K$ is the string stiffness, $F_i$ is the fiber tension, and $F_\tau$ opposes the twisting moment $\tau_L$ \cite{gaponov_twisted_2013}. If the string is assumed to be massless, $F_x$ is simply the force exerted by the load, $F_{\text{load}}$. For very low-stretch materials like ultra-high-molecular-weight polyethylene, the string stiffness is treated as infinite, simplifying to $L_c = L - r_w\phi_\text{eff}$.

\subsection{Velocity Control Law}
To simplify the control laws for the motor velocities, the desired string velocity $\dot{X}_{\text{des}}$ is expressed as the sum of velocities from pure twisting $\dot{X}_\theta$ and pure winching $\dot{X}_\phi$:

\begin{equation}
\dot{X}_{\text{des}} = \dot{X}_\theta + \dot{X}_\phi
\end{equation}
Equation (\ref{eq:linearcontractiontotal}) is differentiated once with respect to time. This process is repeated twice, alternatively setting $\theta_{\text{eff}}$ or $\phi_{\text{eff}}$ to zero. The desired effective twisting velocity \cite{gaponov_twisted_2013} and winch rotation velocity are: 

\begin{equation}
\label{eq:twistvelocity}
\dot{\theta}_{\text{eff}} = \frac{\dot{X}_\theta \sqrt{L_c^2 - \theta_{\text{eff}}^2 r_{\text{var,eff}}^2}}{\theta_{\text{eff}} r_{\text{var,eff}}^2} - \frac{\theta_{\text{eff}} \dot{r}_{\text{var,eff}}}{r_{\text{var,eff}}}
\end{equation}
\begin{equation}
\label{eq:winchvelocity}
\dot{\phi}_{\text{eff}} = \frac{\dot{X}_\phi}{r_w}
\end{equation}
These equations, along with Equation (\ref{eq:motorvelocities}), capture the interaction between twisting and winching, enabling precise actuator control.

%\subsection{Effects of String Stiffness and Load Forces}
% However, unlike standard TSAs, the string of the TSA-winch actuator may experience more significant acceleration due to the winch. If the string has mass and is accelerating due to the winch, axial force is distributed along the string:

% \begin{equation}
% \label{eq:stringforcedistribute}
% F_x(s) = F_x(0) + \lambda\ddot{X}s
% \end{equation}
% %
% \begin{equation}
% F_x(0) = \frac{\tau_\text{w}}{r_w}, \quad F_x(L) = \frac{\tau_w}{r_w} + \lambda\ddot{X}L = F_{\text{load}}
% \end{equation}
% %
% Here, $s$ is a point along the string, $\lambda$ is the string mass per unit length, $\tau_w$ is the winch torque, $r_w$ is the winch radius, and $L$ is the string length. Equation \ref{eq:stringforcedistribute} can be rewritten as:

% \begin{equation}
% F_x(s) = \frac{F_\text{load}-\frac{\tau_w}{r_w}}{L}s + \frac{\tau_w}{r_w}
% \end{equation}

% In reality, it should be pointed out that the string stiffness is generally treated as infinite when using string made of very low-stretch material such as ultra high molecular weight polyethylene. In this case, $L_c = L - r_w\phi_\text{eff}$. Nevertheless, equation (\ref{eq:forcesandcontractedlength}) is included for thoroughness. 

\subsection{Steady-State Forces}

In a scenario where the end of the string is quasi-static---for instance, when the end of a string is pulling at a load but is effectively stalled---it is useful to know the linear forces outputted by the actuator for given winch torque and twisting rotations. In the case where the output end of the string is fixed, the string does not change in length at all, and $\Delta X = 0$ (refer to Detail A in Fig. \ref{fig:linediagram}). As it twists, though, the string stretches according to its stiffness $K$ and exerts force $F_\text{x,twist}$ according to the following equation:

\begin{equation}
F_\text{x,twist} = K\left(\sqrt{X^2 + \theta^2r_0^2}-X\right)\text{sin}(\alpha)
\label{eq:forceztwist}
\end{equation}
where $\alpha$ is the helix angle formed during twisting. The string radius stays constant because the length is fixed. Pure winching contributes force $F_\text{x,winch}$:

\begin{equation}
F_\text{x,winch} = \frac{\tau_w}{r_w}\left(1-\text{sin}(\gamma)\mu\right)
\label{eq:forcezwinch}
\end{equation}
Here, $\mu$ is the sliding friction coefficient between the string material and exit bushing material. $\gamma$ is the angle that the string exiting the turret makes with the section of string spanning the pulley to the turret's exit bushing (Fig. \ref{fig:linediagram}). $\gamma$ can also be defined as $\text{tan}(\gamma)=\frac{r_w}{d_\text{winch}}$, with $d_\text{winch}$ being the distance from the winch's center to the exit bushing. The term $(1-\text{sin}(\gamma)\mu)$ adjusts for the friction as the string slides against the exit bushing. Combining Equations (\ref{eq:forceztwist}) and (\ref{eq:forcezwinch}), the total force output is:

\begin{equation}
F_x = F_\text{x,twist} + F_\text{x,winch}
\label{eq:force}
\end{equation}

\section{EXPERIMENTAL SETUP}

\subsection{General Setup}

The experimental setup used two motors to manipulate a Dacron polyester bowstring, chosen for its high stiffness and unbraided structure. One motor (motor 2 in Fig. \ref{fig:crosssection}) drove a turret through a 2:1 gear reduction transmission to twist the string about its axis. The other (motor 1 in Fig. \ref{fig:crosssection}) rotated a drive shaft that passed through the gear transmission to actuate the winch through a 2:1 reduction bevel gear transmission mounted on the turret. The string, fastened to a flange on the winch, was guided through a low-friction bushing at the turret's tip.

For displacement measurement, the setup used a linear potentiometer and modular microcontroller hardware developed by TinkerForge. The string exiting the turret was clamped to a linear carriage that slid along a rail when the string was twisted or winched. This carriage was linked to an OPH Series linear potentiometer from P3 America that ran parallel to the rail. Because the potentiometer slide lacked bearings and fit loosely in the rail grooves of the potentiometer, flexure mounts constrained it to minimize the binding friction that is common with stages spanning two parallel rigid rails without bearings. A TinkerForge Dual Analog In bricklet read the potentiometer output, with a 20V Powerstack Dewalt battery supplying power to reduce the switching noise affecting voltage readings.

\subsection{Displacement Measurement Setup}

\begin{figure}
    \centering
    \includegraphics[width = 0.99 \linewidth]{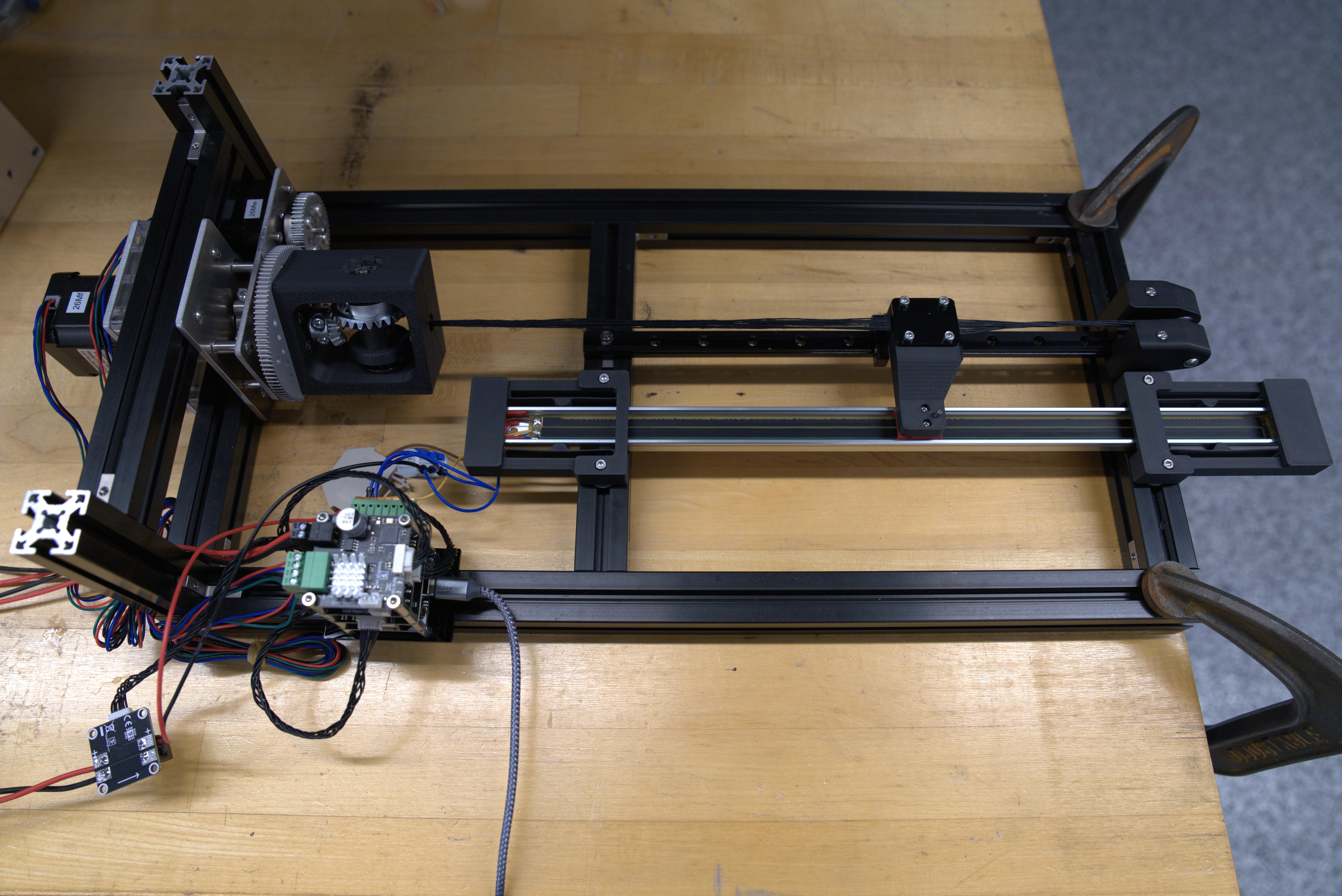}
    \caption{Experimental setup for displacement and velocity tests.}
    \label{fig:experimentalsetup}
\end{figure}

From the clamp on the linear stage, the experimental setup had two forms. To measure displacement, the string passed from the stage to a pulley and then over the table's edge to a hanging set of masses. This removed air gaps in the string and simulated various load conditions (Fig. \ref{fig:experimentalsetup}). Two NEMA 17 stepper motors were chosen for their ease of installation and reliable operation without feedback. Each motor was driven by a TinkerForge Silent Stepper bricklet and powered by a 24V supply. 

\subsection{Force Measurement Setup}

\begin{figure}
    \centering
    \includegraphics[width = 0.99 \linewidth]{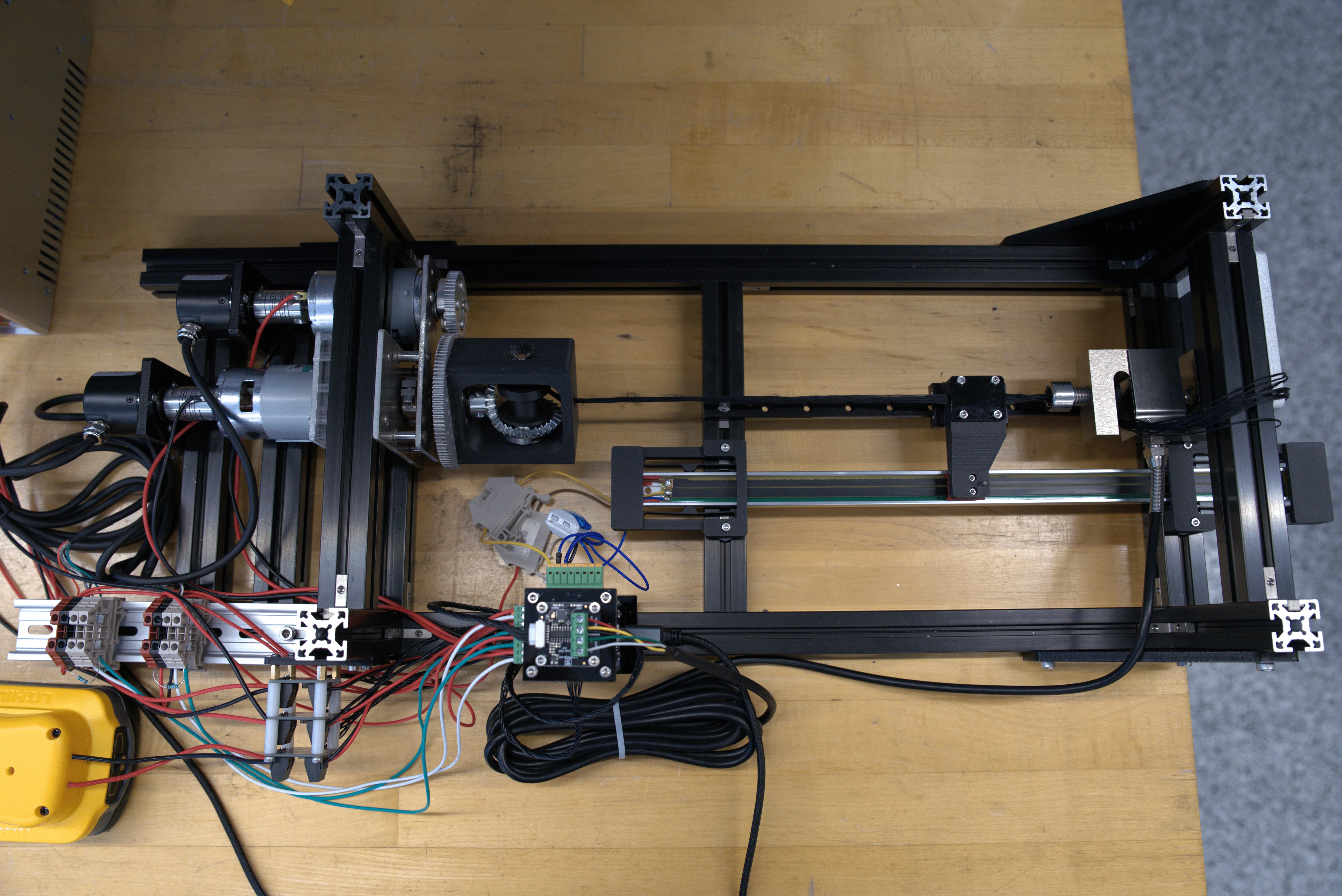}
    \caption{Experimental setup for measuring force.}
    \label{fig:experimentalsetupforces}
\end{figure}

For measuring torque input and force output of this actuator, the experimental setup differed (Fig. \ref{fig:experimentalsetupforces}). Instead of passing from the stage to hanging masses, the string connected the stage to a PSD-S1 type load cell rated for 100 kg via a vented screw. The load cell's signals were sent to a TinkerForge Load Cell bricklet. DC 775 dual shaft brushed motors replaced the stepper motors for more stable current readings by TinkerForge Voltage/Current bricklets. 600 PPR incremental quadrature rotary encoders mounted on the motors’ backs were wired to a TinkerForge Industrial Counter bricklet for use in a PID velocity controller. The torque-current curve of the motors was determined experimentally.     

\section{PERFORMANCE ANALYSIS}

\subsection{Verifying Displacement Model for Different Lengths}

To mimic typical use cases, experiments were conducted, and observed results were compared with model predictions. The most likely applications for this actuator are tendon-driven robots and exoskeletons, particularly for linkages or limbs that must travel a large distance before exerting high forces on a contact surface. Thus, our displacement tests began with pure winching action to achieve a large stroke, followed by twisting for the final high-force motion.

\begin{figure}[t]
    \centering
    \includegraphics[width = 0.99 \linewidth]{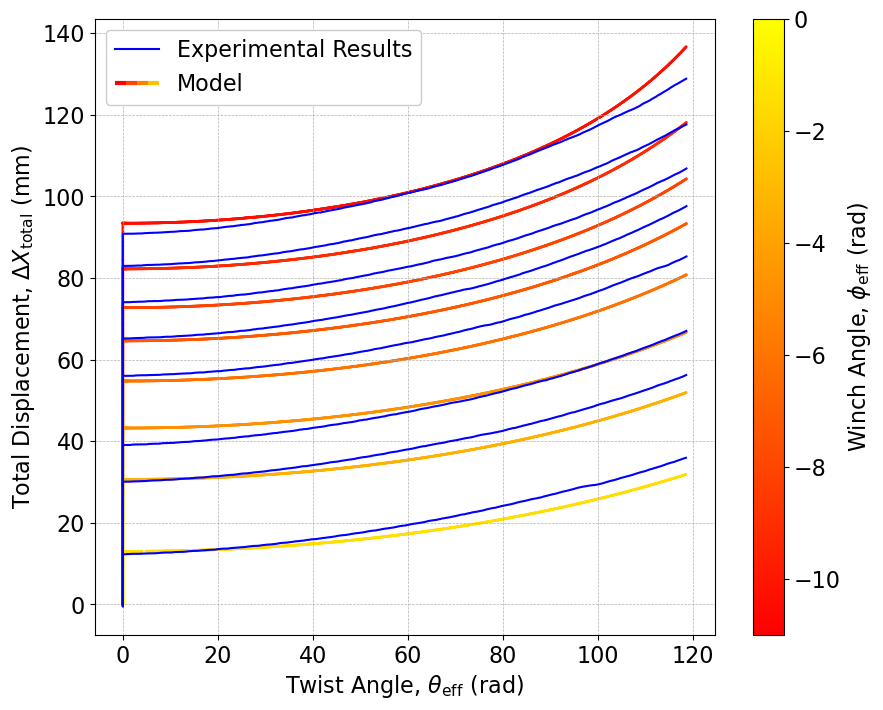}
    \caption{Total linear displacement $\Delta X_{\text{total}}$ with respect to effective twist angle $\theta_{\text{eff}}$ (x-axis) and effective winch rotation $\phi_{\text{eff}}$ (color gradient). The gradient line is the model, and the solid blue line is experimental results. Here, the string was winched to different starting positions in gradually decreasing increments and then twisted from each starting position. The plot shows how varying $\phi_{\text{eff}}$ affects displacement output across different  $\theta_{\text{eff}}$ values.} 
    \label{fig:displacementmultiple}
\end{figure}

\begin{figure}[t]
    \centering
    \includegraphics[width = 0.99 \linewidth]{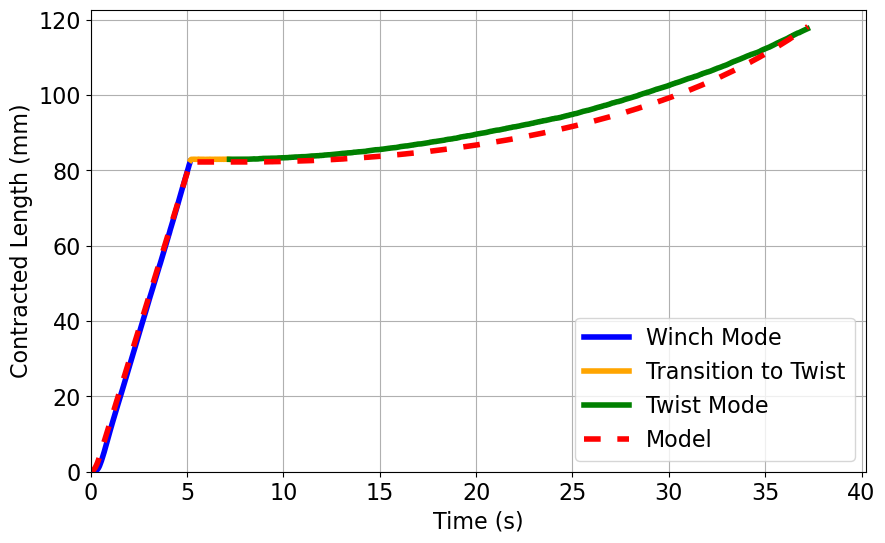}
    \caption{Zoomed-in view of a single winch-twist process. The actuator winds up the string, brakes, and then twists to exert higher force. Experimental contracted length is compared to the model (red dotted line).}
    \label{fig:displacementzoomed}
\end{figure}

Fig. \ref{fig:displacementmultiple} shows the most probable use case: using the winch to move to different positions, then twisting the string to generate high forces. Fig. \ref{fig:displacementzoomed} provides a closer look at a single winch-twisting cycle, with each step segmented and labeled. As expected, winch action dominates displacement. Once twisting begins, displacement appears exponential relative to the twisting angle. Additionally, Fig. \ref{fig:displacementmultiple} shows that displacement due to pure twisting increases as the starting contracted length decreases if the number of twists remains constant, as expected from Equation (\ref{eq:linearcontractiontotal}). From the data, the displacement of the linear stage was 23.76 mm after twisting when 12.17 mm of the string was wound up, whereas the displacement was 38.22 mm after twisting when 90.56 mm of the string was wound up. 

The average percent error and absolute error are 4.47\% and 2.48 mm respectively, demonstrating good adherence of experimental results to the model. Errors in winching could arise from the string fastening point being off-center from the turret's exit bushing. However, the largest errors, as high as 5.88 mm, seem to occur during the twisting phase, where the model doesn't appear to capture the behavior as well. This could be due to the string not fully untwisting before the next winch-twist cycle. Hence, the string starts the next cycle already slightly twisted, as suggested by the steeper start to each experimental twisting curve. Other twisting errors may come from string stretch or poor packing of the strands.

\subsection{Verifying Velocity Model}

\begin{figure}[t]
    \centering
    \includegraphics[width = 0.99 \linewidth]{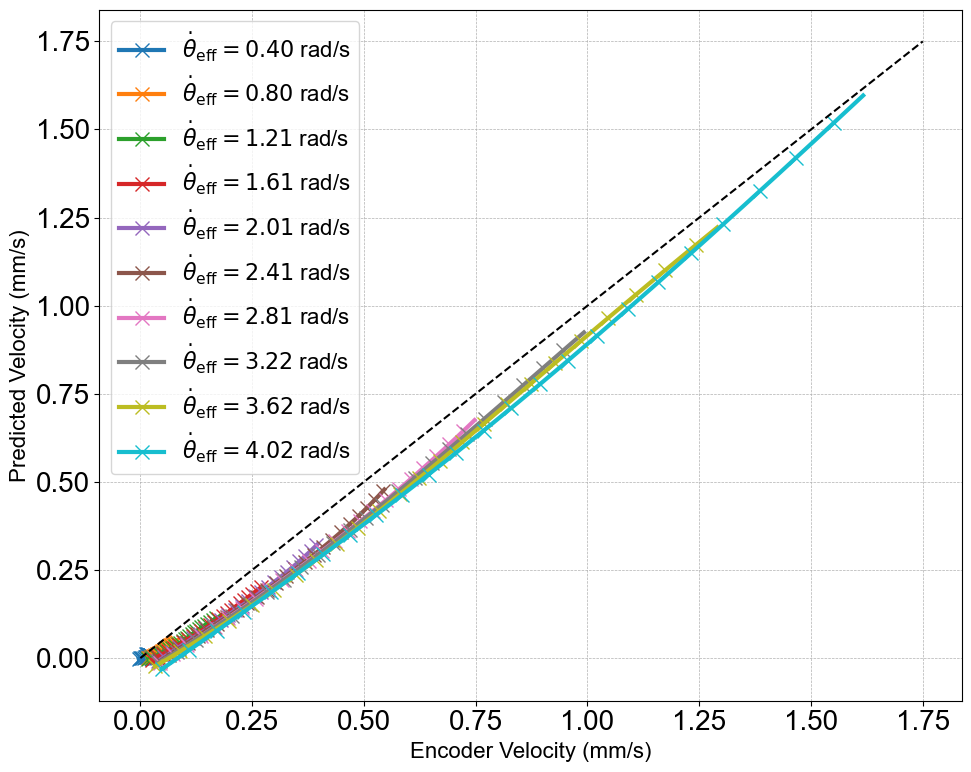}
    \caption{Predicted vs. encoder velocity for effective twist rates \(\dot{\theta}_{\text{eff}}\) from 0.40 to 4.02 rad/s, showing good model fit.}
    \label{fig:velocity1}
\end{figure}

\begin{figure}[t]
    \centering
    \includegraphics[width = 0.99 \linewidth]{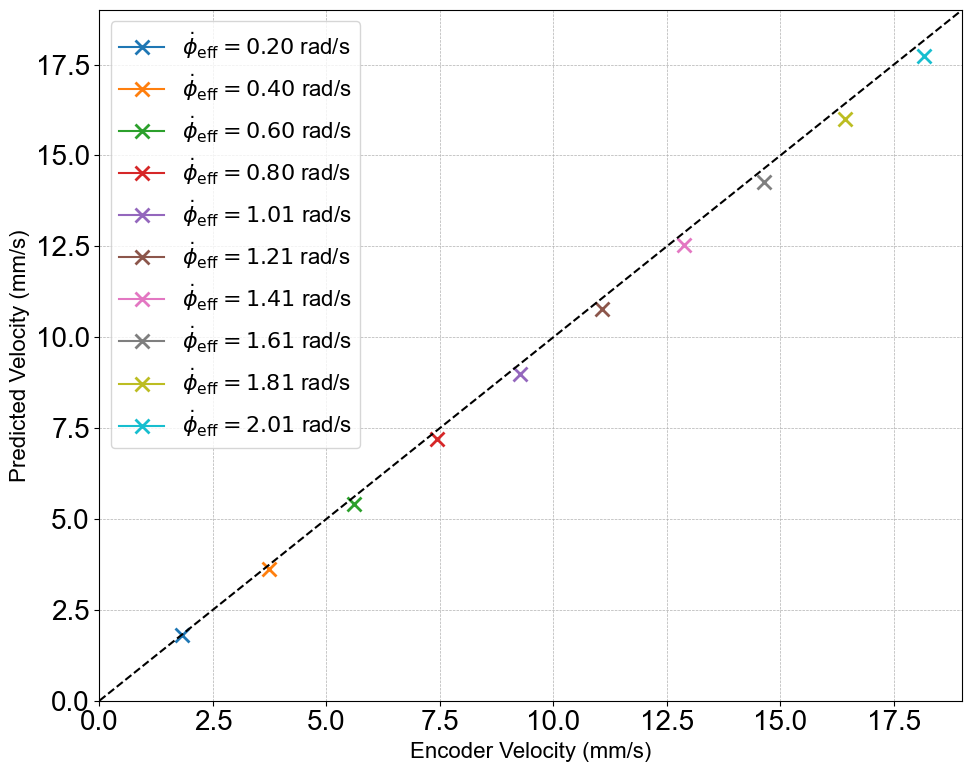}
    \caption{Predicted vs. encoder velocity for higher winch rates \(\dot{\phi}_{\text{eff}}\) from 0.20 to 2.01 rad/s, confirming model accuracy.}
    \label{fig:velocity2}
    
\end{figure}

To validate the velocity model from Equations (\ref{eq:twistvelocity}) and (\ref{eq:winchvelocity}), predicted velocities were compared to measured encoder velocities for various effective twist rates \(\dot{\theta}_{\text{eff}}\). Fig. \ref{fig:velocity1} shows the predicted velocity versus encoder velocity for effective twist rates ranging from 0.40 rad/s to 4.02 rad/s. Despite the maximum error being 0.12 mm/s, the low average absolute error of 0.067 mm/s indicates a good fit between the model and the experimental data, demonstrating the model's ability to predict actuator behavior accurately across twist velocities. Slight deviations from linearity may arise from the same errors noted in the displacement analysis.

Similarly, Fig. \ref{fig:velocity2} shows a near-linear relationship between predicted and encoder velocities for winch rates from 0.20 to 2.01 rad/s, confirming that the model accurately captures the effects of both twisting and winching on the actuator’s performance across different velocities. As expected, the winch achieves much higher velocities than twisting, limited only by motor capabilities. The error between experimental and predicted values further supports this: the maximum error was 0.44 mm/s and the mean error was 0.27 mm/s. These small errors validate the model's reliability in predicting actuator behavior.
%2.60\% percent error)

\subsection{Verifying Force Model}

\begin{figure}[t]
    \centering
    \includegraphics[width = 0.99 \linewidth]{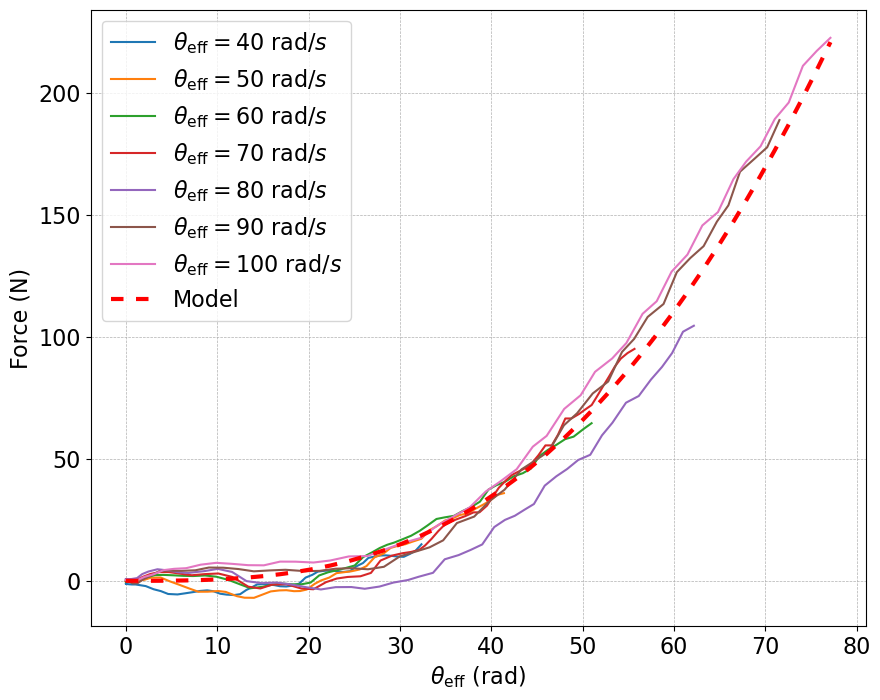}
    \caption{Force vs. effective twist angle \(\theta_{\text{eff}}\) at different twist rates. Solid lines show experimental data for twist rates from 40 to 100 rad/s. The dashed line represents the model prediction, closely matching the experimental results.}
    \label{fig:forcetwist}
\end{figure}

\begin{figure}[t]
    \centering
    \includegraphics[width = 0.99 \linewidth]{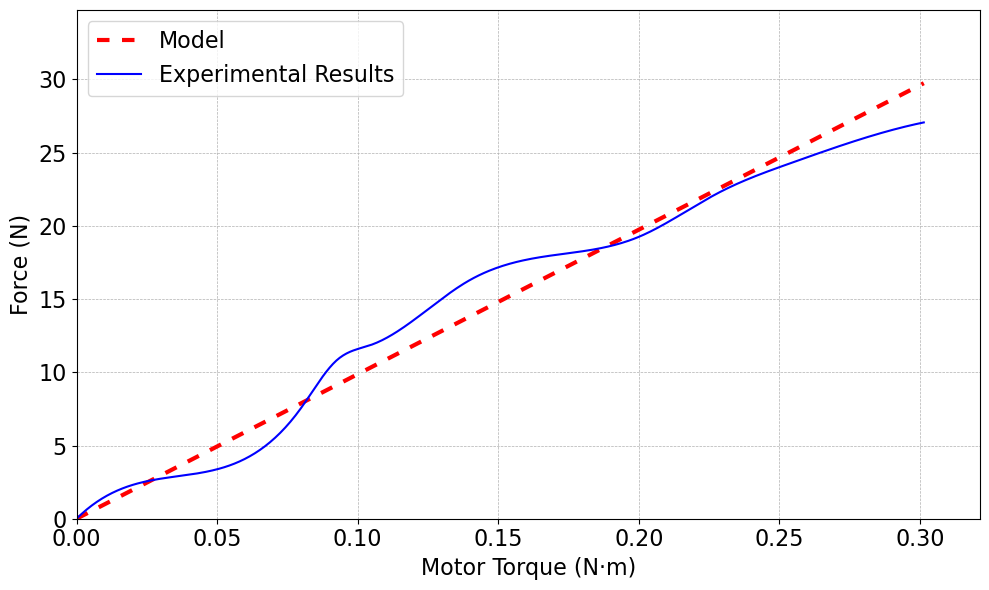}
    \caption{Relationship between the torque applied to the winch and the force exerted at the end of the string. As seen in Equation \ref{eq:force}, this relationship is linear and quite trivial but nonetheless conveys that the winch is poorly equipped to deliver high force.}
    \label{fig:forcewinch}
\end{figure}

The winch enables rapid windup compared to a standard TSA, but the hybrid actuator also exerts high forces through twisting. To demonstrate this, the setup in Fig. \ref{fig:experimentalsetupforces} was used, rotating the turret at various velocities while recording motor current and load cell force readings. For each test, the vented set screw fastened to the load cell was tightened until it measured 2 kg, removing air gaps and preventing string slack through pretension. Tests were limited to one second; longer tests caused the high forces from the twisted string to collapse the setup.

Fig. \ref{fig:forcetwist} shows the results compared to the model based on Equation (\ref{eq:forceztwist}). The string stiffness used in the model was taken from previous work \cite{gaponov_twisted_2013}. Torque readings were consistent across tests: an initial spike to about 0.12 N$\cdot$m to overcome static friction, a dip as the turret reached a constant velocity, then a gradual increase to a peak of 0.16 N$\cdot$m as the motor met the resistance of the stretching string. The average percent error was 3.84\%, and the average absolute error was 4.18 N, demonstrating high adherence to the model expectations. Nevertheless, the maximum error seen was 18.46 N, which could be due to excessive untwisting. The consistent, almost periodic oscillations in the experimental curves suggest slightly misaligned gear meshing, causing more friction at certain points of the rotation. 

To compare twisting and winching for high-force tasks, the winch motor was run at several duty cycles, and the torque and force were recorded (Fig. \ref{fig:forcewinch}). The experimental results closely matched the linear contribution of winch force to the model in Equation (\ref{eq:force}). Though the highest error recorded was 2.78 N, the average percent and absolute errors were 8.06\% and 1.27 N, most likely due to noise from the load cell and noise in the motor current readings. Comparing Figs. \ref{fig:forcetwist} and \ref{fig:forcewinch} shows that with equal or less motor torque, twisting generates much more force depending on the number of twists. If the number of twists was not limited by the compliance of the test setup, the force from twisting would most likely have been shown to be much higher. Overall, these final tests emphasize the dual capabilities of this actuator for both rapid displacement and high force.

\section{CONCLUSIONS AND FUTURE WORK}

This paper demonstrates the effectiveness of combining a twisted string actuator (TSA) with a winch mechanism to overcome the limitations of TSAs, particularly in stroke length and force transmission. The integrated system successfully achieves variable transmission ratios and enhances stroke flexibility, avoiding the issue of overtwisting to achieve large displacement. The mathematical models for displacement, velocity control, and force transmission were validated through experiments, confirming the system's performance. The TSA-winch actuator provides a wide range of transmission ratios and precise control, making it a versatile option for robotic applications. 

This actuator design offers a potentially compact, adaptable solution that has several practical implications for robotics and automation, particularly the wide range of tasks that require both rapid displacement and high forces. For example, a tendon-driven robotic hand could use this actuator to rapidly grasp an object and then squeeze the object to secure it tightly. Another application would be using cables to safely tumble heavy components in an assembly warehouse space, twisting to overcome large friction but winching to efficiently move the large objects.
% or exoskeletons
% Though experiments were done with an unbraided string that characteristically behaves according to the variable radius twisting model,

\subsection{Future Work}
% Future work: dynamic, more thorough study on the effects of the bushing/friction of string around pulley, explore pulley friction when twist then windup, miniaturizing the system, stroke is limited by winch size (too much winding and string will come in contact with bevel gears), modeling braided string vs unbraided, when twisting and then winching some of the twist is stored on the pulley so not unwinding to same point
%"This is most likely caused by the fact that the existing theoretical relations treat the string as a cylinder composed of a number of independent fibers or strands, whereas the braided strings behave differently due to the nature of their internal structure." "Because of such a structure, braided strings form small bulges during twisting [see Fig. 4(c)], and such a behavior is not described by the existing mathematical model" - \cite{gaponov_twisted_2013}

\begin{enumerate}
    \item \textbf{Friction Analysis:} A detailed study of the friction effects, especially around the pulley and bushing, on performance during twist and windup will help optimize the system.

    % \item \textbf{Pulley Friction During Twist and Windup:} Investigating the friction introduced during the twist and windup process will be crucial for refining the system’s efficiency and reliability.

    \item \textbf{Miniaturization:} Future efforts will focus on miniaturizing the actuator while maintaining its performance, enabling its use in smaller robotic systems. 
    %This would include reducing the number of components with more robust machined parts to resist the high forces during the twisting phase.

    \item \textbf{Stroke Limitation:} The current stroke is limited by the winch size, as excessive winding risks contact between the string and bevel gears. Addressing this limitation through improved winch designs or alternative string routing will be a priority.

    \item \textbf{Alternative Strings:} Also worth exploring would be incorporating alternatives to unbraided strings. This would open up the design space for this actuator by allowing the use of other flexible elements such as braided strings made of stiffer material and ribbons with greater mechanical amplification than regular strings.
    
\end{enumerate}
The TSA-winch combination offers significant advantages in stroke flexibility and force transmission. Further research will refine the design, address the identified limitations, and explore new applications for this innovative actuator.

%                                   % the textheight of the last page by a suitable amount.
%                                   % This command does not take effect until the next page
%                                   % so it should come on the page before the last. Make
%                                   % sure that you do not shorten the textheight too much.

%%%%%%%%%%%%%%%%%%%%%%%%%%%%%%%%%%%%%%%%%%%%%%%%%%%%%%%%%%%%%%%%%%%%%%%%%%%%%%%%

%%%%%%%%%%%%%%%%%%%%%%%%%%%%%%%%%%%%%%%%%%%%%%%%%%%%%%%%%%%%%%%%%%%%%%%%%%%%%%%%

%%%%%%%%%%%%%%%%%%%%%%%%%%%%%%%%%%%%%%%%%%%%%%%%%%%%%%%%%%%%%%%%%%%%%%%%%%%%%%%%
%\section*{APPENDIX}


%\section*{ACKNOWLEDGMENT}

%The preferred spelling of the word ÒacknowledgmentÓ in America is without an ÒeÓ after the ÒgÓ. Avoid the stilted expression, ÒOne of us (R. B. G.) thanks . . .Ó  Instead, try ÒR. B. G. thanksÓ. Put sponsor acknowledgments in the unnumbered footnote on the first page.

%%%%%%%%%%%%%%%%%%%%%%%%%%%%%%%%%%%%%%%%%%%%%%%%%%%%%%%%%%%%%%%%%%%%%%%%%%%%%%%%

\bibliographystyle{IEEEtran}
\bibliography{zoterotwistwinchreferences.bib}

% \end{thebibliography}

\end{document}